\documentclass[letterpaper]{article}
\usepackage{aaai22}  
\usepackage{times}  
\usepackage{helvet}  
\usepackage{courier}  
\usepackage[hyphens]{url}  
\usepackage{graphicx} 
\usepackage{natbib}  
\usepackage{caption} 
\DeclareCaptionStyle{ruled}{labelfont=normalfont,labelsep=colon,strut=off} 
\usepackage{booktabs}       
\usepackage{tikz}           
\usepackage{subcaption}
\usepackage{microtype}
\usepackage{indentfirst}
\usetikzlibrary{shapes.geometric, arrows}

\tikzstyle{startstop} = [rectangle, minimum width=2cm, minimum height=1cm, rounded corners, text centered, draw=black, fill=gray!30]
\tikzstyle{process} = [rectangle, minimum width=2cm, minimum height=1cm, text centered, draw=black]
\tikzstyle{arrow} = [thick,->,>=stealth]

\begin{document}

\title{Post-OCR Document Correction with large Ensembles of Character Sequence-to-Sequence Models}

\author{Juan Ramirez-Orta\textsuperscript{\rm 1}\thanks{Corresponding author. Please send correspondence to juan.ramirez.orta@dal.ca}, Eduardo Xamena\textsuperscript{\rm 2}, Ana Maguitman\textsuperscript{\rm 3, 4}, Evangelos Milios\textsuperscript{\rm 1}, Axel J. Soto\textsuperscript{\rm 3, 4}}
    
\affiliations{
     \textsuperscript{\rm 1}Department of Computer Science, Dalhousie University\\
     \textsuperscript{\rm 2}Institute of Research in Social Sciences and Humanities (ICSOH), Universidad Nacional de Salta - CONICET\\
     \textsuperscript{\rm 3}Department of Computer Science and Engineering, Universidad Nacional del Sur\\
     \textsuperscript{\rm 4}Institute for Computer Science and Engineering, UNS--CONICET 
}

\maketitle

\begin{abstract}
   In this paper, we propose a novel method to extend sequence-to-sequence models to accurately process sequences much longer than the ones used during training while being sample- and resource-efficient, supported by thorough experimentation. To investigate the effectiveness of our method, we apply it to the task of correcting documents already processed with Optical Character Recognition (OCR) systems using sequence-to-sequence models based on characters. We test our method on nine languages of the ICDAR 2019 competition on post-OCR text correction and achieve a new state-of-the-art performance in five of them. The strategy with the best performance involves splitting the input document in character n-grams and combining their individual corrections into the final output using a voting scheme that is equivalent to an ensemble of a large number of sequence models. We further investigate how to weigh the contributions from each one of the members of this ensemble. Our code for post-OCR correction is shared at \url{https://github.com/jarobyte91/post_ocr_correction}.
\end{abstract}

\section{Introduction}

Since its inception in the early sixties, OCR has been a promising and active area of research. Nowadays, systems like Tesseract \cite{tesseract} obtain accuracies above 90\% on documents from 19th- and early 20th-century newspaper pages \cite{annual_test}, but the accurate recognition of older, historical texts remains an open challenge due to their vocabulary, page layout, and typography. This is why successful OCR systems are language-specific and focus only on resource-rich languages, like English.

As a consequence of these difficulties, the task of automatically detect and correct errors in documents has been studied for several decades \cite{kukich1992techniques}, ranging from techniques based on statistical language modelling \cite{tong1996statistical}, dictionary-based translation models \cite{kolak2002ocr} or large collections of terms and word sequences \cite{bassil2012ocr}.

With the advent of methods based on neural networks, and more specifically, sequence models such as \cite{cho_seq2seq,sutskever_seq2seq,transformer}, the automatic correction of texts using sequence models witnessed considerable progress in the form of neural sequence models based on characters or words \cite{neubig_ocr_correction,schnober-etal-2016-still}. 

Character-based sequence models offer good generalization due to the flexibility of their vocabulary, but they are challenging to train and inefficient at inference time, as generating a document one character at a time requires thousands of steps. On the other hand, word-based sequence models are efficient at inference time and more sample-efficient than character-based sequence models, but they lack generalization, a problem that has been partially solved with systems like WordPiece \cite{wordpiece} or Byte-Pair Encodings \cite{bytepair}, that learn useful sub-word units to represent text from the data they are trained on.

In this work, we propose a novel method to correct documents of arbitrary length based on character sequence models. The novelty of our method lies in training a character sequence model on short windows both to detect the mistakes and to generate the candidate corrections at the same time, instead of first finding the mistakes and then use a dictionary or language model to correct them, as is usual with post-OCR text correction systems. 

The first main idea behind our method is to use the sequence model to correct n-grams of the document instead of the whole document as a single sequence. In this way, the document can be processed efficiently because the n-grams are corrected in parallel. The other key idea of the method is the combination of all the n-gram corrections into a single output, a process that adds robustness to the technique and is equivalent to using an ensemble of a large number of sequence models, where each one acts on a different segment.

The features that set apart the method proposed in this paper from previous methods for post-OCR text correction are the following:
\begin{itemize}
    \item It can handle documents of great length and difficulty while being character-based, which means that it can deal with out-of-vocabulary sequences gracefully and be easily applied to various languages.
    \item It is sample- and resource-efficient, requiring only a couple of hundred corrected documents in some cases to produce good improvements in the quality of the text while needing very modest hardware to train and to perform inference.
    \item It is robust because it integrates a set of strategies to combine the output of a large ensemble of character sequence models, each one focusing on a different context.
    \item It sets a new state-of-the-art performance on the ICDAR 2019 competition for post-OCR text correction. The system hereby proposed obtained major improvements in Spanish, German, Dutch, Bulgarian and Czech, while remaining competitive in the remaining languages.
\end{itemize}

\section{Related Work}

The state of the art in OCR post-processing is reflected in the two editions of the ICDAR competition on Post-OCR text correction \cite{chiron2017icdar2017,rigaud2019icdar}. This competition is divided into two tasks: the detection of OCR errors and their correction. 

The best performing error detector method during the first edition of the challenge was WFST-PostOCR \cite{wfst}, while the best correction method was Char-SMT/NMT \cite{char-smt}. WFST-PostOCR relies on compiling probabilistic character error models into weighted finite-state edit transducers, while a language model finds the best token sequence. On the other hand, Char-SMT/NMT is based on ensembles of character-based Machine Translation models, each one trained on texts from different periods of time to translate each token within a window of two preceding and one succeeding tokens. 

In the second edition of the challenge, the best method for both error detection and correction was Context-based Character Correction (CCC). This method is a fine-tuning of multilingual BERT \cite{devlin2018bert} that applies a machine translation technique based on a character sequence model with an attention mechanism.  

The most recent extension to the CCC method also applies BERT and character-level machine translation \cite{nguyen2020neural}, but it also includes static word embeddings and character embeddings used in a Neural Machine Translation system, and a candidate filter. The method proposed in \cite{schaefer2020two} argues that applying a two-step approach to automatic OCR post-correction reduces both the Character Error Rate (CER) and the proportion of correct characters that were falsely changed. The resulting model consists of a bidirectional LSTM-based detector and a standard LSTM-based sequence-to-sequence translation model.

Unlike CCC, our method does not rely on pretrained language models, which makes it applicable to low-resource settings without sacrificing performance.
\section{Methodology}

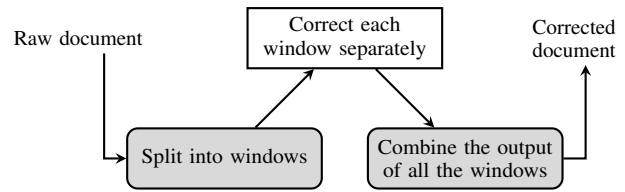
\begin{figure} [htb]
    \centering
    \begin{tikzpicture} [thick,scale=0.8, every node/.style={transform shape}, node distance = 2cm, text width = 3cm] 
        \node (document) {Raw document};
        \node (split) [right of = document, below of = document] [startstop] {Split into windows};
        \node (correct) [above of = split, right of = split] [process] {Correct each window separately};
        \node (combine) [below of = correct, right of = correct] [startstop] {Combine the output of all the windows};
        \node (improved) [above of = combine, right of = combine] {\hspace{0.5cm} Corrected \\\hspace{0.5cm} document};
        \draw [arrow] (document) |- (split);
        \draw [arrow] (split) -- (correct);
        \draw [arrow] (correct) -- (combine);
        \draw [arrow] (combine) -| (improved);
    \end{tikzpicture}
    \caption{Overview of the proposed method. In the first step, the document is split into either disjoint windows or n-grams. In the second step, the windows are corrected in parallel using the sequence model. In the third step, the partial corrections obtained in the previous step are combined to obtain the final output: by a simple concatenation when using disjoint windows or a voting scheme when using n-grams. After the merging step, the final output can be compared with the correct transcription using Character Error Rate.}
    \label{fig:overview}
\end{figure}

The main idea of our method is to train a sequence model on sequences of characters and then use it to correct complete documents. However, using this approach directly is computationally unfeasible because documents are sequences of thousands of characters, and training a model like this would need an immense amount of both memory and corrected documents. To overcome these limitations, we propose a method composed of three steps, as shown in Fig. \ref{fig:overview}.

\subsection{The Sequence Model}

The core of our system is a standard sequence-to-sequence model that can correct sequences of characters. In our implementation, we used a Transformer \cite{transformer} as the sequence model, which takes as input a segment of characters from the document to correct, and the output is the corrected segment. To train this sequence model, it is necessary to align the raw documents with their corresponding correct transcriptions, which is not always straightforward.

Since the output is not necessarily of the same length as the input (because of possible insertions or deletions of characters), a decoding method like Greedy Search or Beam Search is needed to produce the most likely corrected sequence according to the model.

\subsection{Processing Full Documents with the Sequence Model at Inference Time}

Assuming that the sequence model is already trained, the next step is to use it to correct texts of arbitrary length. This can be done by splitting the document into windows with a length similar to the ones on which the model was trained and combining them with the strategies we describe next.

\subsubsection{Disjoint Windows}

Correcting a document by splitting it into disjoint windows is the most basic way to use the sequence model to process a string that is longer than the maximum sequence it allows. In the splitting step, the string to correct is split into disjoint windows of a fixed length $n$. In the correction step, each window is corrected in parallel using the sequence model. In the merging step, the final output is produced by concatenating the corrected output from each window. To evaluate the method, the final output can be compared with the correct transcription using CER. 

\begin{figure}[htb]
    \centering
    \includegraphics[width=\columnwidth]{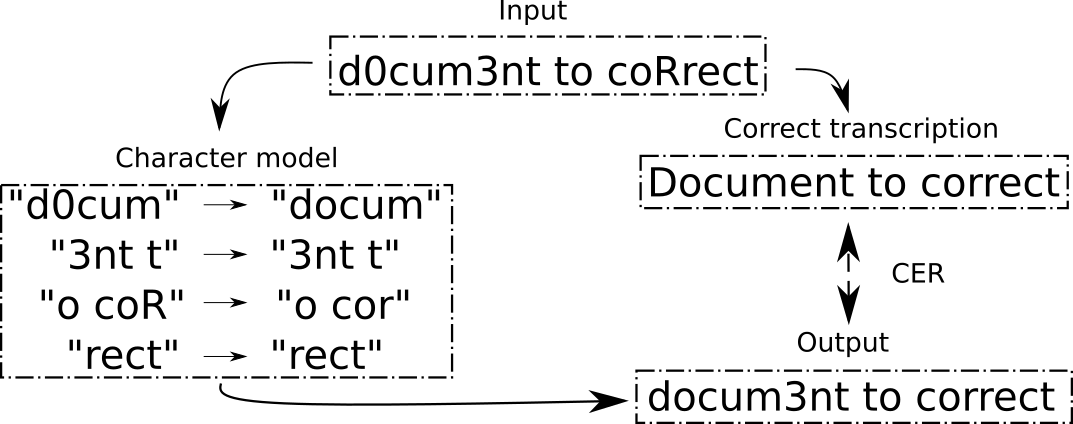}
    \caption{An example of correcting a document using disjoint windows of length 5.}

    \label{fig:disjoint}
\end{figure}

It is important to note that this approach can be effective if the sequence model is well trained, but if this is not the case, it can be prone to a ``boundary effect", where the characters at the ends of the windows do not have the appropriate context. An example of this approach is shown in Fig. \ref{fig:disjoint}.

\subsubsection{N-Grams}

\begin{figure}[htb]
    \centering
    \includegraphics[width=\columnwidth]{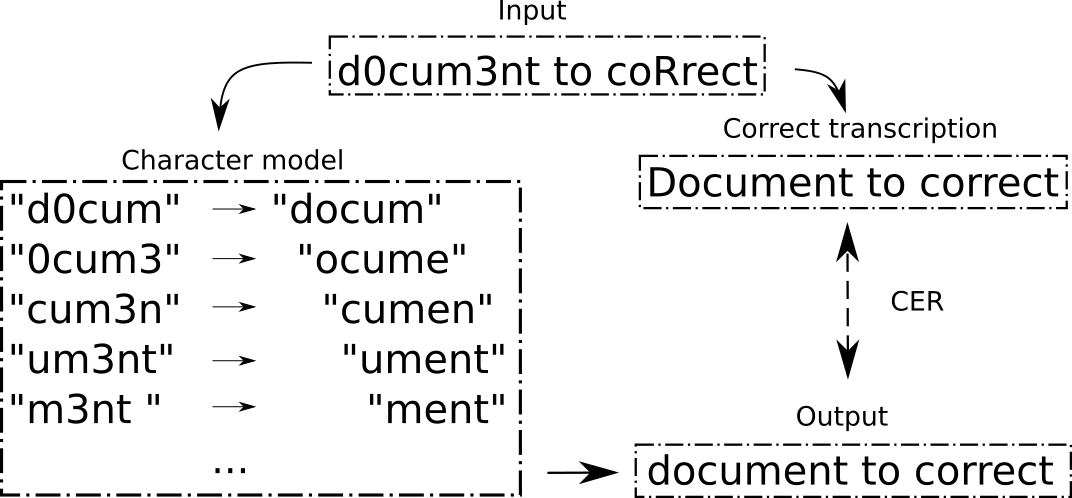}
    \caption{An example of correcting a document using n-grams of length 5.}
    \label{fig:n-grams}
\end{figure}

To counter the ``boundary effect", it is possible to add robustness to the output by using all the n-grams of the input. In the splitting step, the string to correct is split into character n-grams. In the correction step, each window is corrected in parallel using the sequence model. The merging step produces the final output by combining the output from the windows, taking advantage of the overlapping between them and a voting scheme influenced by a weighting function described below. To evaluate the method, the final output is compared with the correct transcription using CER. An example of this method is depicted in Fig. \ref{fig:n-grams}.

An essential part of the n-grams variation is how the partial outputs are combined. Since the partial corrections have an offset of one, the outputs can be combined by aligning them and performing a vote to obtain the most likely character for every position. This vote is equivalent to processing the whole input with an ensemble of $n$ models, each one operating on segments of offset 1, where $n$ is the order of the n-grams. 

Since a character corrected in the middle of an n-gram has more context than a character in the edges, it is reasonable to think that they should have different weights in the vote. To express this difference, we used three different weighting functions, given by the following formulae:

$$
\mbox{\textit{bell}}(p, w) = exp\left(-\left(1 - \frac{p}{m}\right)^2\right),
$$

$$
\mbox{\textit{triangle}}(p, w) = 1 -\frac{\vert m - p \vert}{2m},
$$

$$
\mbox{\textit{uniform}}(p, w) = 1,
$$
where $p$ is the character position in the window, $w$ is the window length, and $m = \lceil \frac{w}{2} \rceil$. 

\begin{figure}[htb]
    \centering
    \includegraphics[width=0.8\columnwidth]{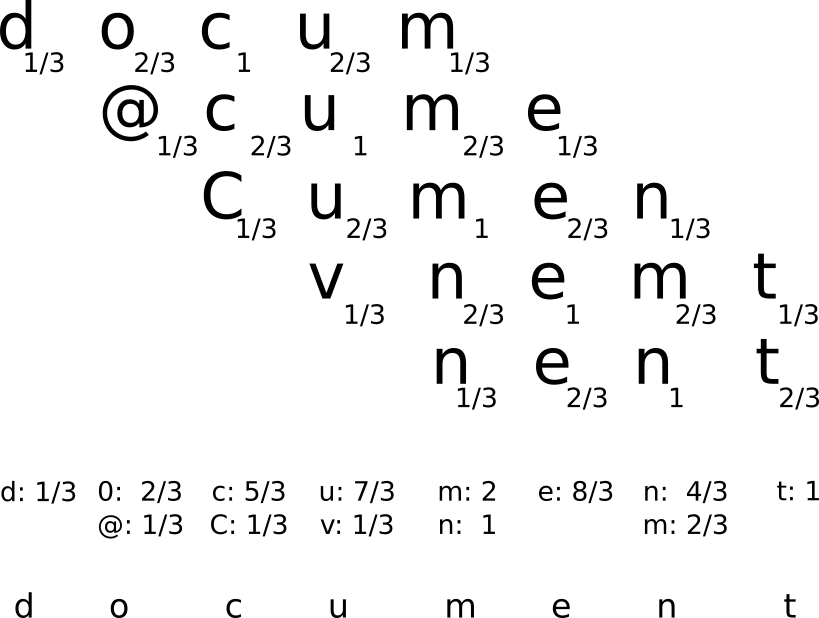}
    \caption{An example of correcting a text with 5-grams and the \textit{triangle} weighting function. The number under every character in the top part is the weight of that character in its position for every window. The mid-bottom table shows the sum of the weights for every candidate character on each position of the output. To generate the final output (at the bottom), the candidate character with the maximum sum on every position is selected.}
    \label{fig:weighting}
\end{figure}

The weight of the character vote in position $p$ in an n-gram of length $w$ is given by $f(p, w)$, where $f$ is one of the weighting functions. An example of this is shown in Fig. \ref{fig:weighting}.

\section{Experimental Setup}

\subsection{Data}

The dataset of the ICDAR2019 Competition on Post-OCR Text Correction is made of 14,309 documents scanned with OCR along with their corresponding correct transcription in 10 languages: Bulgarian (bg), Czech (cz), German (de), English (en), Spanish (es), Finnish (fi), French (fr), Dutch (nl), Polish (pl) and Slovak (sl). In this work, we used all the languages except Finnish because the files required are distributed separately due to copyright reasons. The details of the datasets used are shown in Table \ref{table:icdar}.

\begin{table}[htb]
\centering
\resizebox{\columnwidth}{!}{
\begin{tabular}{lrrrrrr}
\toprule
Language & \multicolumn{1}{l}{Total} & $\mu$ length & $\mu$ CER & $\sigma$ CER & \multicolumn{1}{l}{Train} & \multicolumn{1}{l}{Best $\%$}  \\
 & documents & & & & documents & improvement \\\midrule
bg       & 198    & 2,332       & 16.65 & 16.30  & 149    & 9.0                \\
cz       & 195    & 1,650       & 5.99  & 12.98  & 149    & 6.0               \\
de       & 10,080 & 1,546       & 24.57 & 5.86   & 8,052  & 24.0               \\
en       & 196    & 1,389       & 22.76 & 23.81  & 148    & 11.0               \\
es       & 197    & 2,876       & 31.52 & 22.65  & 147    & 11.0               \\
fr       & 2,849  & 1,521       & 8.79  & 12.15  & 2,257  & 26.0               \\
nl       & 198    & 4,289       & 28.11 & 25.00  & 149    & 12.0               \\
pl       & 199    & 1,688       & 36.68 & 20.50  & 149    & 17.0               \\
sl       & 197    & 1,538       & 12.50 & 19.85  & 149    & 14.0               \\ \bottomrule
\end{tabular}%
}
\caption{The ICDAR datasets. $\mu$ length is the average document length measured in characters. $\mu$ CER and $\mu$ CER are the mean and standard deviation of the Character Error Rate between every document and its correct transcription. Best $\%$ improvement is the percentage of improvement in the Character Error Rate from the best method reported in \protect \cite{rigaud2019icdar}.}
\label{table:icdar}
\end{table}


\subsection{Obtaining Sequence Pairs for the Sequence Model}

\begin{figure}[htb]
    \centering
    \includegraphics[width=\columnwidth]{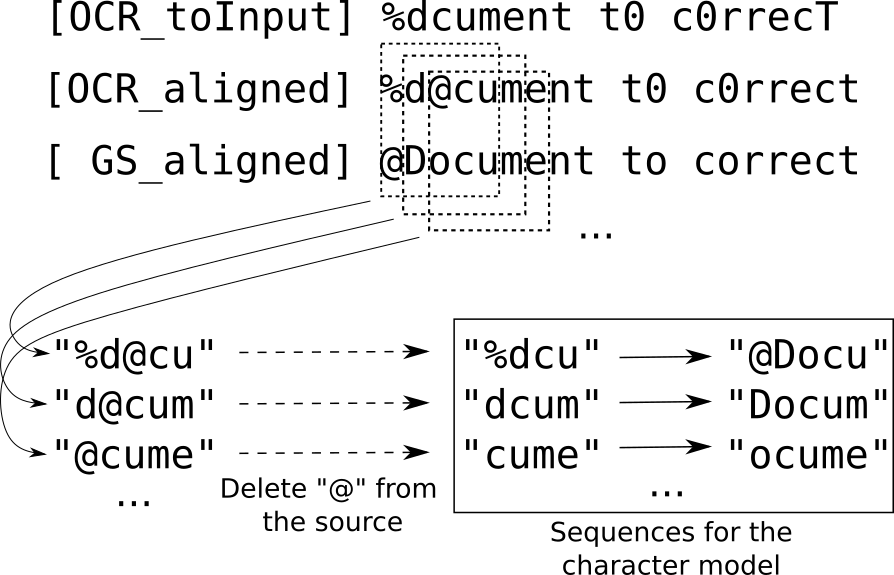}
    \caption{An example of the process to train the sequence model using the ICDAR datasets with windows of length 5. In the first step, the correct transcription of the document ($GS\_aligned$) is split into n-grams, and for each one, the corresponding part of the aligned input ($OCR\_aligned$) is retrieved. In the second step, the padding character ``@" is deleted only from the aligned input to obtain a set of segments from the document ($OCR\_toInput$) paired with their corresponding correction.}
    \label{fig:obtaining_sequences}
\end{figure}

To obtain the character sequences to train the sequence model, the format of the ICDAR datasets was crucial. The alignment process we followed is described in Fig. \ref{fig:obtaining_sequences}.

To create a development set for each language, we sampled five documents from each training set and then split the ground truth of every document into n-grams of length 100 to create the input-correction pairs to train and develop the sequence models. We chose this number of documents to be able to evaluate the models frequently and this length because this was the largest one that fitted in our hardware with the largest architectures we tried. The datasets used to train our models are described in Table \ref{table:datasets}.

\begin{table}[htb]
\centering
\resizebox{\columnwidth}{!}{
\begin{tabular}{@{}l|rrr|rrr@{}}
\toprule
Language &  \multicolumn{3}{c|}{Train} & \multicolumn{3}{c}{Development}        \\ 
         & $\mu$ length & $\mu$ CER & Pairs      & $\mu$ length & $\mu$ CER       & Pairs  \\ \midrule
bg       & 1,872    & 16.14 & 278.3    & 1,708    & 9.39        & 8.7   \\
cz       & 1,638    & 6.02  & 238.3    & 2,017    & 10.56       & 10.1  \\
de       & 1,547    & 24.52 & 12,779.5 & 1,531    & 22.84       & 7.7   \\
en       & 1,419    & 23.83 & 217.9    & 1,295    & 45.62       & 7.3   \\
es       & 2,967    & 30.84 & 466.2    & 2,110    & 43.40       & 11.4  \\
fr       & 1,534    & 8.63  & 3,553.8  & 1,643    & 5.53        & 4.9   \\
nl       & 4,293    & 28.38 & 666.6    & 3,762    & 32.62       & 21.5  \\
pl       & 1,666    & 40.08 & 259.6    & 1,463    & 29.95       & 7.8   \\
sl       & 1,383    & 11.24 & 208.2    & 1,457    & 1.25        & 7.3   \\ \bottomrule
\end{tabular}%
}
\caption{Datasets used to train the sequence models. $\mu$ length is the average character length of the documents. $\mu$ CER is the average Character Error Rate between each document and its correct transcription. Pairs is the number in thousands of segment-correction pairs obtained.}
\label{table:datasets}
\end{table}

\subsection{Training the Sequence Models}

The process of training the models is the standard sequence-to-sequence pipeline that uses cross entropy loss to make the model generate the right token at every step, as it was originally proposed in \cite{cho_seq2seq, sutskever_seq2seq}. All the models were trained using 4 CPU cores, 4 GB of RAM, and a single GPU NVIDIA V100 with 16 GB of memory. Overall, training the sequence models was difficult because of the differences between the training and development sets, but the models obtained were good enough to produce improvements in all the languages, as shown in Table \ref{table:models}.

\begin{table}[htb]
\centering
\resizebox{\columnwidth}{!}{

\begin{tabular}{@{}lrrrrrrrrr@{}}
\toprule
Language & \multicolumn{1}{l}{Best}  & \multicolumn{1}{l}{Total}  & \multicolumn{1}{l}{Best}     & \multicolumn{1}{l}{Best}       & \multicolumn{1}{l}{Parameters} & Training \\ 
         & \multicolumn{1}{l}{epoch} & \multicolumn{1}{l}{epochs} & \multicolumn{1}{l}{dev loss} & \multicolumn{1}{l}{train loss} & \multicolumn{1}{l}{}           & hours    \\ \midrule
bg       & 19                        & 42                         & 0.278                        & 0.251                          & 1.94                           & 2.19     \\
cz       & 2                         & 50                         & 0.255                        & 0.095                          & 15.05                          & 3.65     \\
de       & 7                         & 7                          & 0.330                         & 0.406                          & 2.00                          & 1.93      \\
en       & 25                        & 50                         & 1.010                        & 0.455                          & 3.84                           & 1.52     \\
es       & 19                        & 24                         & 1.077                        & 0.688                          & 3.86                           & 1.61     \\
fr       & 10                        & 12                         & 0.318                        & 0.288                          & 1.48                           & 1.88     \\
nl       & 8                         & 16                         & 0.583                        & 0.468                          & 7.54                           & 2.97     \\
pl       & 10                        & 47                         & 0.594                        & 0.578                          & 7.56                           & 3.41     \\
sl       & 15                        & 57                         & 0.035                        & 0.157                          & 3.82                           & 1.78     \\ \bottomrule
\end{tabular}%
}
\caption{Training of the sequence models. ``Best epoch" is the epoch with the lowest development loss. ``Best dev loss" is the lowest loss obtained on the development set. ``Best train loss" is the loss on the training set in the best epoch. ``Parameters" is the number of model parameters in millions.}
\label{table:models}
\end{table}

To tune the hyperparameters of the sequence models, we performed a Random Search \cite{random_search}. We set the embedding dimension to be 128, 256, or 512, with the number of hidden units in the feedforward layers always four times the embedding dimension. We tried from two to four layers, with the same number of layers for both the encoder and the decoder. We varied the dropout rate from 0.1 to 0.5 in steps of 0.1 and the $\lambda$ of the weight decay $L^2$ penalization to be $10^{-1}$, $10^{-2}$, $10^{-3}$ or $10^{-4}$. All the models were trained with Adam and a learning rate of $10^{-4}$. The best hyperparameters found are shown in Table \ref{table:hyperparameters}.

\begin{table}[htb]
\centering
\resizebox{\columnwidth}{!}{
\begin{tabular}{@{}lrrrrr@{}}
\toprule
Language & Embedding & Feedforward & Layers & Dropout \\ 
         & dimension & dimension   &        &         \\ \midrule
bg       & 128       & 512         & 4      & 0.2     \\
cz       & 512       & 2,048        & 2      & 0.1     \\
de       & 128       & 512         & 4      & 0.3     \\
en       & 256       & 1,024        & 2      & 0.5     \\
es       & 256       & 1,024        & 2      & 0.4     \\
fr       & 128       & 512         & 3      & 0.5     \\
nl       & 256       & 1,024        & 4      & 0.2     \\
pl       & 256       & 1,024        & 4      & 0.3     \\
sl       & 256       & 1,024        & 2      & 0.2     \\ \bottomrule
\end{tabular}%
}
\caption{Hyperparameters of the sequence models. All the models were trained with Adam and a learning rate of $10^{-4}$ with a weight decay $L^{2}$ penalization of $10^{-4}$.}
\label{table:hyperparameters}
\end{table}

\subsection{Experimental Results}

\begin{table*}[ht]
\centering
    \resizebox{\textwidth}{!}{%
\begin{tabular}{@{}llrllrrrrr@{}}
\toprule
Language & Window   & Window & Decoding  & Weighting & Inference & $\mu$ CER & $\mu$ CER & \% Improvement & \% Baseline \\ 
         &  type    & size   & method    &           & minutes   & before    & after   &    \\
 \midrule
bg       &  n-grams &           80 &     beam &   uniform &             198.08 &       18.23 &      15.27 &        \textbf{16.27} &       9.0 \\
cz       &  n-grams &           40 &     beam &   uniform &              37.90 &        5.90 &       4.52 &        \textbf{23.36} &       6.0 \\
de       &  n-grams &          100 &     beam &  triangle &            4,340.23 &       24.77 &      15.62 &       \textbf{36.94} &      24.0 \\
en       &  n-grams &           20 &     beam &   uniform &              10.37 &       19.47 &      18.00 &         7.52 &      11.0 \\
es       &  n-grams &           60 &     beam &  triangle &              70.58 &       33.54 &      29.41 &        \textbf{12.30} &      11.0 \\
fr       &  n-grams &           90 &     beam &  triangle &             889.27 &        9.40 &       7.88 &        16.18 &      26.0 \\
nl       &  n-grams &           80 &   greedy &   uniform &              47.35 &       27.30 &      22.41 &        \textbf{17.94} &      12.0 \\
pl       &  n-grams &           10 &   greedy &   uniform &               1.68 &       26.56 &      23.19 &        12.69 &      17.0 \\
sl       &  n-grams &           90 &     beam &   uniform &              85.73 &       16.42 &      14.64 &        10.83 &      14.0 \\
\midrule
& & & & & Average & 20.17 & 16.77 & 17.11 & 14.4\\
\bottomrule
\end{tabular}%
}
\caption{Best approach found for every language on the ICDAR test sets. ``$\mu$ CER before" and ``$\mu$ CER after" is the average Character Error Rate between every document and its correct transcription, before and after using our method. ``\% Improvement" is the average percentage of improvement in CER. ``\% Baseline" is the average percentage of improvement in CER from the best method in \cite{rigaud2019icdar}. The percentage of improvement is bolded when it is larger than the baseline.}
\label{table:best}
\end{table*}

To investigate the effect of the different hyperparameters in our method, we performed a Grid Search varying the window size from 10 to 100 in steps of 10, processing the documents with disjoint windows or n-grams with all the weighting functions using both Greedy Search and Beam Search. 

The best model found for each language is shown in Table \ref{table:best}. The effect of each one of the hyperparameters (window type, decoding method, weighting function and window size) in the average improvement in CER is shown in Tables \ref{table:window type}, \ref{table:decoding}, \ref{table:weighting} and  \ref{table:window size}. The best improvement in CER obtained for every combination of language and window size is shown in Table \ref{table:language window size}. The average percentage of improvement in CER for every combination of language, window type, decoding method, and weighting function is shown in Table \ref{table:language variation improvement}. The average inference time in minutes for every combination of language, window type, decoding method, and weighting function is shown in Table \ref{table:inference times}.

\begin{table}[htb]
\centering
\resizebox{\columnwidth}{!}{%
\begin{tabular}{lrrrrrrrr}
\toprule
Window type   & Count & Mean  & Std   & Min     & 25\%  & 50\% & 75\%  & Max   \\ 
\midrule
disjoint &  180.0 & -6.83 &  65.48 & -422.21 & -2.21 &   4.12 &  11.63 &  36.12 \\
n-grams  &  540.0 &  0.11 &  67.31 & -423.77 &  6.10 &  10.82 &  16.67 &  36.94 \\
\bottomrule
\end{tabular}%
}
\caption{Descriptive statistics of the average percentage of improvement in CER on the ICDAR test sets grouped by window type.}
\label{table:window type}
\end{table}

\begin{table}[htb]
\centering
\resizebox{\columnwidth}{!}{%
\begin{tabular}{@{}lrrrrrrrr@{}}
\toprule
Decoding method & Count & Mean  & Std   & Min     & 25\%  & 50\% & 75\%  & Max   \\ 
\midrule
beam     &  360.0 & -6.33 &  79.12 & -423.77 &  3.69 &  9.74 &  16.07 &  36.94 \\
greedy   &  360.0 &  3.09 &  51.51 & -403.78 &  5.48 &  9.03 &  16.06 &  35.20 \\
\bottomrule
\end{tabular}%
}
\caption{Descriptive statistics of the average percentage of improvement in CER on the ICDAR test sets grouped by decoding method.}
\label{table:decoding}
\end{table}

\begin{table}[htb]
\centering
\resizebox{\columnwidth}{!}{%
\begin{tabular}{@{}lrrrrrrrr@{}}
\toprule

Weighting & Count & Mean  & Std   & Min     & 25\%  & 50\% & 75\%  & Max   \\ \midrule
bell      &  180.0 & -0.10 &  67.47 & -423.76 &  5.94 &  10.43 &  16.36 &  36.89 \\
triangle  &  180.0 &  0.03 &  67.50 & -423.77 &  6.06 &  10.56 &  16.58 &  36.94 \\
uniform   &  180.0 &  0.41 &  67.32 & -423.76 &  6.38 &  10.92 &  16.77 &  36.83 \\
\bottomrule
\end{tabular}%
}
\caption{Descriptive statistics  of the average percentage of improvement in CER  on the ICDAR test sets grouped by weighting function.}
\label{table:weighting}
\end{table}

\begin{table}[htb]
\centering
\resizebox{\columnwidth}{!}{%
\begin{tabular}{@{}rrrrrrrrr@{}}
\toprule
Window & Count & Mean  & Std   & Min     & 25\%  & 50\% & 75\%  & Max   \\ 
size & & & & & & & & \\\midrule
10          &   72.0 & -36.93 &  132.72 & -423.73 &  2.27 &   5.39 &  12.48 &  31.70 \\
20          &   72.0 & -25.77 &  112.54 & -423.77 &  4.74 &   8.00 &  14.58 &  33.22 \\
30          &   72.0 & -16.15 &   96.76 & -408.21 &  4.64 &   8.64 &  15.48 &  33.79 \\
40          &   72.0 &   3.47 &   34.78 & -156.62 &  5.36 &   8.45 &  16.55 &  34.59 \\
50          &   72.0 &  11.37 &   11.59 &  -21.38 &  6.10 &  10.00 &  17.08 &  36.12 \\
60          &   72.0 &  11.72 &   11.33 &  -25.74 &  6.53 &  11.96 &  17.15 &  36.16 \\
70          &   72.0 &  12.43 &   10.97 &  -21.27 &  6.68 &  12.46 &  16.30 &  36.19 \\
80          &   72.0 &  11.89 &   12.34 &  -29.57 &  6.61 &  11.78 &  16.82 &  36.56 \\
90          &   72.0 &   8.06 &   15.71 &  -47.30 & -1.39 &  10.41 &  15.75 &  36.63 \\
100         &   72.0 &   3.70 &   26.46 &  -93.51 & -7.47 &   9.01 &  16.74 &  36.94 \\
\bottomrule
\end{tabular}%
}
\caption{Descriptive statistics  of the average percentage of improvement in CER  on the ICDAR test sets grouped by window size.}
\label{table:window size}
\end{table}

\begin{table*}[!htb]
\centering
\resizebox{0.8\textwidth}{!}{%
\begin{tabular}{@{}l|rrrrrrrrrr@{}}
\toprule
Language & \multicolumn{10}{c}{Window size}  \\ 
 & 10          & 20      & 30     & 40    & 50    & 60    & 70    & 80    & 90    & 100   \\ 
\midrule
bg       & -366.79 & -229.28 & -63.67 &   4.40 &  13.01 &  14.60 &  16.02 &  \textbf{16.27} &  15.77 &  15.42 \\
cz       &   16.66 &   20.80 &  21.81 &  \textbf{23.36} &  22.33 &  18.49 &  19.76 &  21.61 &  15.84 &  22.02 \\
de       &   31.70 &   33.22 &  33.79 &  34.59 &  36.12 &  36.16 &  36.19 &  36.56 &  36.63 &  \textbf{36.94} \\
en       &    5.45 &    \textbf{7.52} &   6.88 &   7.10 &   6.26 &   6.13 &   4.75 &   2.31 &  -1.06 &  -7.03 \\
es       &    4.82 &    8.00 &   9.35 &  11.00 &  12.03 &  \textbf{12.30} &  11.91 &  11.79 &  10.92 &   9.14 \\
fr       &    8.47 &   10.93 &  11.50 &  11.80 &  13.44 &  14.68 &  15.34 &  15.81 &  \textbf{16.18} &  16.07 \\
nl       &   14.35 &   15.94 &  16.50 &  17.10 &  17.45 &  17.49 &  17.82 &  \textbf{17.94} &  17.73 &  17.14 \\
pl       &   \textbf{12.69} &   12.47 &  10.48 &   9.45 &   7.21 &   7.27 &   7.39 &   8.55 &   8.31 &   8.62 \\
sl       &    5.46 &    6.20 &   6.44 &   6.97 &   7.95 &   9.84 &  10.08 &  10.23 &  \textbf{10.83} &   9.24 \\
\bottomrule
\end{tabular}%
}
\caption{Best improvement in CER obtained for every language and for every window size on the ICDAR test sets. The best performance found for every language is bolded.}
\label{table:language window size}
\end{table*}

\begin{table*}[!htb]
\centering
\resizebox{0.8\textwidth}{!}{%
\begin{tabular}{@{}l|rr|rrr|rrr@{}}
\toprule
 Language   & \multicolumn{2}{c|}{disjoint} & \multicolumn{6}{c}{n-grams}                                \\ 
  & beam         & greedy        & \multicolumn{3}{c|}{beam}     & \multicolumn{3}{c}{greedy}  \\
 &           &            & bell    & triangle & uniform & bell   & triangle & uniform \\ 
 \midrule
bg       &  -134.21 & -71.40 & -129.94 &  -129.87 & -129.26 & -61.10 &   -60.93 &  -59.86 \\
cz       &    14.73 &  13.61 &   19.50 &    19.67 &   19.91 &  19.17 &    19.32 &   19.43 \\
de       &    33.13 &  31.21 &   35.11 &    35.13 &   34.97 &  33.33 &    33.36 &   33.20 \\
en       &    -3.06 &  -3.16 &    1.81 &     1.90 &    2.14 &   2.96 &     3.05 &    3.22 \\
es       &     3.37 &   4.58 &    6.75 &     6.79 &    6.90 &   7.71 &     7.73 &    7.75 \\
fr       &     9.61 &   2.03 &   12.68 &    12.84 &   13.39 &  11.14 &    11.32 &   11.93 \\
nl       &     4.54 &   7.10 &   13.89 &    14.02 &   14.52 &  15.99 &    16.10 &   16.41 \\
pl       &   -26.01 &  -1.39 &  -12.21 &   -11.92 &  -10.66 &   8.24 &     8.51 &    9.24 \\
sl       &    -2.12 &  -5.50 &    7.74 &     7.92 &    8.30 &   5.41 &     5.53 &    5.94 \\
\bottomrule
\end{tabular}%
}
\caption{Average percentage of improvement of CER by language for each variation of our method on the ICDAR test sets.}
\label{table:language variation improvement}
\end{table*}

\begin{table*}[!htb]
\centering
\resizebox{0.8\textwidth}{!}{%
\begin{tabular}{@{}l|rr|rrr|rrr@{}}
\toprule
 Language   & \multicolumn{2}{c|}{disjoint} & \multicolumn{6}{c}{n-grams}                                \\ 
  & beam         & greedy        & \multicolumn{3}{c|}{beam}     & \multicolumn{3}{c}{greedy}  \\
 &           &            & bell    & triangle & uniform & bell   & triangle & uniform \\ 
 \midrule
bg       &     3.42 &   0.84 &   269.11 &   270.16 &   275.63 &   38.73 &    38.55 &   38.11 \\
cz       &     1.88 &   0.49 &   160.87 &   161.15 &   160.77 &   30.49 &    30.54 &   30.54 \\
de       &    61.77 &  22.78 &  4,489.84 &  4,340.23 &  4,372.81 &  606.48 &   602.05 &  612.37 \\
en       &     0.93 &   0.39 &    66.70 &    66.29 &    64.41 &   10.51 &    10.53 &   10.55 \\
es       &     1.79 &   0.46 &   149.66 &   149.99 &   148.28 &   23.23 &    23.31 &   23.31 \\
fr       &    13.38 &   6.14 &  1,617.90 &   934.24 &   932.64 &  127.17 &   127.33 &  127.40 \\
nl       &     4.53 &   1.03 &   443.59 &   422.70 &   424.34 &   70.29 &    70.38 &   69.87 \\
pl       &     2.03 &   0.72 &   175.27 &   172.24 &   166.71 &   28.53 &    28.50 &   28.48 \\
sl       &     1.30 &   0.42 &   101.95 &   102.48 &   100.76 &   16.27 &    16.35 &   16.30 \\
\bottomrule
\end{tabular}%
}
\caption{Average inference time in minutes for every language and every variation of our method on the ICDAR test sets.}
\label{table:inference times}
\end{table*}
\section{Discussion}

Our method outperformed the state of the art in Bulgarian (bg), Czech (cz), German (de), Spanish (es), and Dutch (nl), while exhibiting comparable performance in the remaining languages, as shown in Table \ref{table:best}. The results obtained are interesting for several reasons:

\begin{itemize}
    \item The method was not as effective in French as it was in German, the other language with abundant training data.
    \item The choice of weighting function did not have much impact on the performance, although broadly speaking, the best weighting function was \textit{uniform}.
    \item Although the method is stable with respect to changes in the window size, a larger window size does not always lead to improved performance. It can sometimes hurt the model's performance, a behavior that appears to be language-dependent, as in the case of English and Polish, according to Table \ref{table:language variation improvement}.
    \item Although the best results were consistently obtained with Beam Search, Greedy Search seems to be a safer choice than Beam Search. Using Beam Search is between three and ten times slower than using Greedy Search, but these extra computations are usually not justified given that there is no guarantee of increased performance, and even when the performance does increase, the difference is small, as shown in Table \ref{table:inference times}.
\end{itemize}

\begin{figure}[htb]
     \begin{center}
     \includegraphics[width=\columnwidth]{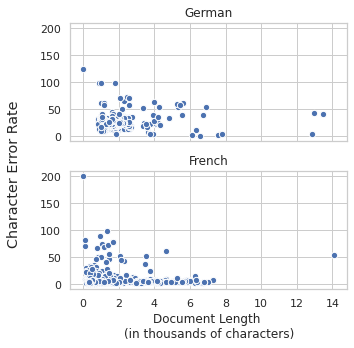}
    \caption{Distribution of the length in characters against the Character Error Rate for each document in the German and French datasets.}
    \label{fig:distributions}
     \end{center}
\end{figure}

It is important to note that the datasets come from several heterogeneous sources with varying levels of quality and content. In the French dataset, we noticed two important properties: a large portion of the documents are receipts, with little to no narrative text, while the longest documents have very few errors, therefore not allowing much room for improvement, as shown in Fig. \ref{fig:distributions}.

After informal manual inspection of the testing sets, we observed that the French model mostly learned to discard parts of the document and to correct numbers and dates. On the other hand, the German model learned to correct the narrative parts.
It is important to also note that most models in the original competition also performed poorly in French, while those with the best performance in French used external resources such as Google Book N-grams \cite{rigaud2019icdar}.

\section{Conclusions and Future Work}

The method proposed in this paper allows processing very long texts using character sequence-to-sequence models, which makes it applicable to any language. The method is simple, resource-efficient and easily parallelizable, obtaining from modest to very good improvements in documents of varying length and difficulty.

Although this paper is focused on text and post-OCR correction, the methods presented here can be transferred to many other sequence problems that require only local dependencies to be solved successfully, requiring very modest hardware and just a couple hundred examples in some cases.

For future work, it would be interesting to apply this method to text from Automated Speech Recognition or Handwritten Text Recognition systems, but the problem of aligning the system's output with the correct transcription remains.

\section{Acknowledgments}

We thank Calcul Quebec (\url{https://www.calculquebec.ca/en/}), Compute Canada (\url{www.computecanada.ca}), Dalhousie University, CONICET (PUE 22920160100056CO) and CIUNSa (Project C 2659) for the resources provided to enable this research. We would also like to thank all the anonymous reviewers of this work for their valuable feedback and suggestions. 

\bibliography{references}

\end{document}